\documentclass[letterpaper, 10 pt, conference]{ieeeconf}  

\IEEEoverridecommandlockouts                              
\usepackage[utf8]{inputenc}
\usepackage{amsmath,amssymb,stmaryrd,mathtools}
\usepackage{amsfonts}
\usepackage[flushleft]{threeparttable}
\usepackage{acronym}
\usepackage{verbatim}
\usepackage{booktabs}
\usepackage{siunitx}
\usepackage{graphics}
\usepackage{graphicx,caption}
\usepackage[keeplastbox]{flushend}
\usepackage{suffix}
\usepackage{xstring}
\usepackage{xparse}
\usepackage{expl3}
\usepackage{mathrsfs}
\usepackage{tabularx}
\usepackage{makecell}
\usepackage{array}
\usepackage{hyperref}
\usepackage{cleveref}
\usepackage{multirow}
\usepackage{diagbox}
\usepackage{rotating}
\usepackage{algorithm}
\usepackage[noend]{algpseudocode}

\usepackage{subcaption}
\usepackage{wrapfig}

\usepackage{tikz}
\usetikzlibrary{arrows,backgrounds,calc}
\usepackage{relsize}
\usepackage{float}
\usepackage{kantlipsum} 
\usepackage{lipsum}
\usepackage{stfloats}
\usepackage{siunitx}
\usepackage[noadjust]{cite}

\usepackage{todonotes}
\usepackage{soul}
\definecolor{smoothgreen}{rgb}{0.7,1,0.7}
\sethlcolor{smoothgreen}

\RequirePackage{luatex85}
\usepackage{pgfplots}
\pgfplotsset{compat=newest}
\pgfplotsset{every axis legend/.append style={%
		cells={anchor=west}}
}
\usepgfplotslibrary{polar}
\usetikzlibrary{arrows}
\tikzset{>=stealth'}

\definecolor{C1}{rgb}{0.0, 0.447, 0.741}
\definecolor{C1_light}{rgb}{0.0, 0.6032388663967612, 1.0}
\definecolor{C2}{rgb}{0.85, 0.325, 0.098}
\definecolor{C3}{rgb}{0.929, 0.694, 0.125}
\definecolor{C4}{rgb}{0.494, 0.184, 0.556}
\definecolor{C5}{rgb}{0.466, 0.674, 0.188}
\definecolor{C6}{rgb}{0.301, 0.745, 0.933}
\definecolor{C7}{rgb}{0.635, 0.078, 0.184}

\usepgfplotslibrary{groupplots}

\usetikzlibrary{shapes.geometric, arrows}

\tikzstyle{startstop} = [rectangle, rounded corners, minimum width=2cm, minimum height=1cm,text centered, draw=black, fill=none]
\tikzstyle{arrow} = [thick,->,>=stealth]

\title{
Learning Visual-Audio Representations for Voice-Controlled Robots
}

\author{Peixin Chang, Shuijing Liu,  D. Livingston McPherson, and Katherine Driggs-Campbell
\thanks{P. Chang, S. Liu, D. L. McPherson, and K. Driggs-Campbell are with the Department of  Electrical and Computer Engineering at the University of Illinois at Urbana-Champaign. emails: \{pchang17,sliu105,dlivm,krdc\}@illinois.edu}%
\thanks{This work is supported by Agriculture and Food Research Initiative (AFRI) grant no. 2020-67021-32799/project accession no.1024178 from the USDA National Institute of Food and Agriculture.}%
}

\begin{document}

\maketitle
\thispagestyle{empty}
\pagestyle{empty}

\begin{abstract}
Based on the recent advancements in representation learning, we propose a novel pipeline for task-oriented voice-controlled robots with raw sensor inputs. Previous methods rely on a large number of labels and task-specific reward functions. Not only can such an approach hardly be improved after the deployment, but also has limited generalization across robotic platforms and tasks. To address these problems, our pipeline first learns a visual-audio representation (VAR) that associates images and sound commands. 
Then the robot learns to fulfill the sound command via reinforcement learning using the reward generated by the VAR.
We demonstrate our approach with various sound types, robots, and tasks.
We show that our method outperforms previous work with much fewer labels. We show in both the simulated and real-world experiments that the system can self-improve in previously unseen scenarios given a reasonable number of newly labeled data.
\end{abstract}

\section{Introduction}
\label{sec:intro}
\begin{figure}[t]
    \centering
    \vspace{5pt}
    \includegraphics[scale=0.65]{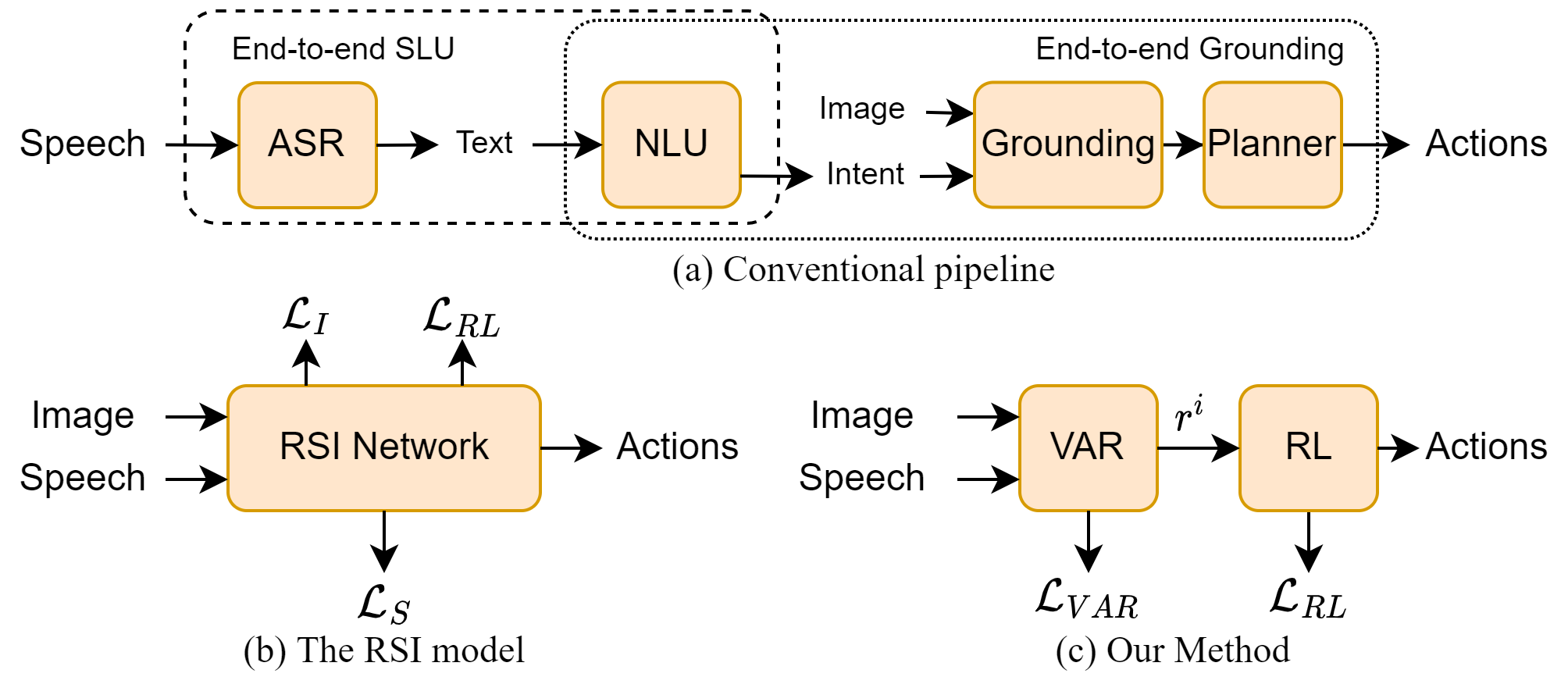}
    \caption{\textbf{Illustration of different models for voice-controlled robots.} (a) A usual pipeline for voice-controlled robots.  Independent modules process the inputs. End-to-end SLU combines the ASR and the NLU, while end-to-end grounding combines the NLU and the grounding module. (b) The first RSI model~\cite{chang2020robot}. Auxiliary losses such as $\mathcal{L}_{I}$ and $\mathcal{L}_{S}$ are used for classifying images and the intent in the speech, respectively. The model is trained end-to-end with an RL loss, $\mathcal{L}_{RL}$. (c) The proposed model. The intrinsic reward is denoted by $r^i$.}
    \label{fig:diff_model}
    \vspace{-20pt}
\end{figure}

As robots begin to enter people's daily lives, there is an increasing need for non-experts to intuitively interact and communicate with robots, which leads to a wealth of research in voice-controlled robots~\cite{liu2019review}. Conventional voice-controlled robot consists of independent modules for automatic speech recognition (ASR), natural language understanding (NLU), symbol grounding, and motion planning (Fig.~\ref{fig:diff_model}a). However, this design suffers from intermediate and cascading errors among different modules and the difficulties in grounding language to the physical entities~\cite{tada2020robust}.
To resolve the problems, end-to-end models directly map text-based commands and image observations to a sequence of robot actions that fulfill the commands~\cite{anderson2018vision,yu2018interactive,hermann2017grounded,chaplot2018gated,Shridhar2020ALFRED}. These models are first trained in a simulated environment using reinforcement learning (RL), imitation learning, and auxiliary losses, and then the models are deployed to a real robot without further training~\cite{vln-pano2real}.
The first end-to-end model for task-oriented voice-controlled robots and Robot Sound Interpretation (RSI) is proposed in~\cite{chang2020robot}, as illustrated in Fig.~\ref{fig:diff_model}b. The RSI agent learns to ground from the raw image and audio signals and can handle both speech and non-speech input. 
 

\begin{figure}[!t]
    \centering
    \includegraphics[scale=0.66]{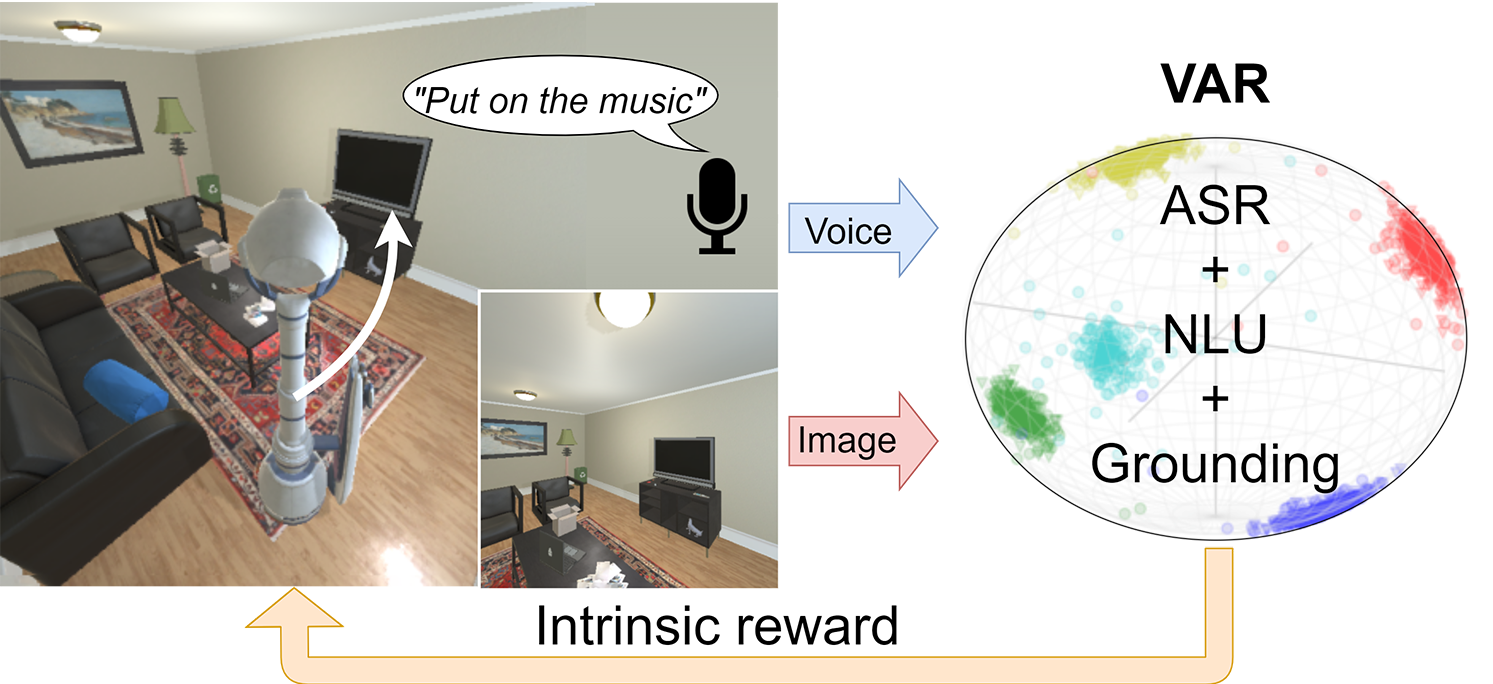}
    \caption{\textbf{Illustration of our method.} 
   The agent navigates and finishes household tasks according to a voice command. Windowed image shows the robot camera view. The voice and image received by the robot are the input to the VAR. The VAR is a unification of ASR, NLU, and grounding and produces an intrinsic reward that supports self-supervised RL training.}
    \label{fig:opening}
    \vspace{-20pt}
\end{figure}

While previous works made noticeable progress, they have some common weaknesses. 
First, the RL-based methods require hand-tuned and task-specific reward functions, which limits the generalization of the methods.
Second, it is cost-prohibitive for non-experts to fine-tune these methods. Ideally, when an intelligent voice-controlled robot encounters unseen scenarios such as new speakers or new room layouts, it should be customizable and can \textit{continually} improve its interpretation of language and skills from non-experts in daily life. However, the performance of the models degrades due to domain shift~\cite{james2019sim,akkaya2019solving,Tobin_2017}. Improving the system requires domain expertise, hardware setup, accurate measurements for reward calculation, a prohibitive number of labels for classification and recognition, or speech transcriptions for ASRs and NLUs, all of which are hard to obtain from and not intuitive to non-experts.

Based on the recent advancements in representation learning, we address these weaknesses by proposing a two-stage pipeline as shown in Fig.~\ref{fig:opening} and Fig.~\ref{fig:diff_model}c. At the first stage, we train a visual-audio representation (VAR) using paired data that can be collected easily and intuitively by non-experts.
At the second stage, we use the VAR to produce an intrinsic reward function for RL training. The intrinsic reward function is general across tasks or robots, which helps avoid laborious reward engineering.
After the robot is deployed in the real world, non-experts can collect a reasonable amount of new data intuitively and update the VAR and the intrinsic reward without providing speech transcriptions, class labels, or tuned reward functions. The robot can then improve its skill in a self-supervised manner. 

We apply this learning approach to different robotic tasks in diverse settings as illustrated in Fig.~\ref{fig:opening} and Fig.~\ref{fig:qualitative_result}. Given a sound command, the robot must identify the commander's goal (intent), draw the correspondence between the raw visual and audio inputs, and develop a policy to finish the task. The tasks are challenging because the observation mainly comes from a monocular uncalibrated RGB camera. No maps, depth images, human demonstrations, or prior knowledge is available.

Our main contributions are as follows. (1) We present a novel voice controlled robot pipeline whose application is not constrained to a specific dataset or benchmark. (2) We build a novel VAR that provides reward signals to the downstream RL training. To our best knowledge, the VAR is the first representation that unifies the ASR, NLU, and the grounding module. (3) Our system exhibits promising results in multiple navigation and manipulation tasks. We demonstrate that our pipeline enables the robot to self-improve with limited amount of data from a new domain using both simulated and real-world experiments. (4) We create and solve the first AI environment that uses state-of-the-art free-form speech commands dataset for instruction-following tasks.




\section{Related Works}
\label{sec:related}
\subsection{Conventional voice-controlled robots}
\label{sec:relatedA}
As shown in Fig.~\ref{fig:diff_model}a, a modular voice-controlled robots usually consists of four independent modules: an ASR system to transcribe speech to text~\cite{fernandez2016natural}, an NLU system to map text transcripts to speaker intent~\cite{bastianelli2016discriminative, tada2020robust}, a grounding module to associate the intent with physical entities~\cite{paul2018efficient, magassouba2019understanding,wang2022audio}, and a planner to generate feasible trajectories for task execution~\cite{stramandinoli2016grounding,chen2020enabling, liu2019review, shridhar2020ingress}. However, the pipeline is prone to intermediate errors. Off-the-shelf ASRs usually operate in a general-purpose setting irrelevant to specific robotic tasks. Typical NLU systems, however, are trained on clean text~\cite{serdyuk2018towards}. As a result, the text outputs could be inevitably out-of-context or erroneous~\cite{vanzo2016robust, tada2020robust}, which causes the overall task performance to degrade. 
Rule-based systems based on symbolic AI are used for grounding~\cite{savage2019semantic}. However, developing such system requires significant labor and these systems do not scale beyond their programmed domains~\cite{tada2020robust, hermann2017grounded}. 

\subsection{End-to-end language understanding and grounding}
End-to-end spoken language understanding (SLU) system maps the speech directly to the speaker’s intent without translating the speech to text~\cite{serdyuk2018towards, lugosch2019speech, kim2021st}. The end-to-end learning avoids the intermediate errors of ASR+NLU pipeline and allows the models to fully exploit subtle information such as speaker emotion that are lost during transcription~\cite{lugosch2019speech, kim2021st}. 
However, end-to-end SLU systems are mainly used for virtual digital assistants and not robotics. 

End-to-end language grounding agents are used to perform tasks according to text-based natural language instructions and visual observations~\cite{anderson2018vision,yu2018interactive,hermann2017grounded,chaplot2018gated,Shridhar2020ALFRED}. Our work is different from them in two aspects. First, we focus on voice-controlled agents, while the above works consider text-based input which limits the potential of natural human-robot communication. Second, training these agents require either expert demonstrations, step-by-step instructions~\cite{anderson2018vision,Shridhar2020ALFRED}, or a carefully designed extrinsic reward function~\cite{yu2018interactive,hermann2017grounded,chaplot2018gated}. In contrast, our method needs none of these above, and thus requires less efforts to improve in a new domain.  

RSI is the first work that combines the ideas from end-to-end SLU and language grounding agents for voice-controlled robots, where the ASR, the NLU, the grounding module, and the robot control are unified as an end-to-end model~\cite{chang2020robot}. 
However, the method in \cite{chang2020robot} has limited ability to improve in different domains and generalize across different tasks and robots, as mentioned in Sec.~\ref{sec:intro}. In contrast, our work can be improved in a new domain and is general across tasks.


\subsection{Representation learning for robotics}
Representation learning has shown great potential in learning useful embeddings for downstream robotic tasks. Deep autoencoders are used to encode high-dimensional observations such as images into lower-dimensional vector  embeddings. These vectors are then used as states or intrinsic reward functions for RL~\cite{nair2018visual, wang2020roll}. At test time, however, the methods in~\cite{nair2018visual, wang2020roll} require users to provide goal images for task execution while our method takes voice commands, which is an more natural and convenient way of human-robot communication. Additionally, reconstruction of the input images makes the autoencoders computationally expensive. Another line of work uses contrastive loss to learn representations for downstream tasks such as grasping and water pouring~\cite{sermanet2018time, jang2018grasp2vec, nguyen2020robot}. Contrastive loss avoids the reconstruction required by autoencoders. Different from all of these works that focus mainly on the visual or the text modality, we address the interplay between sight and sound.  


\begin{figure*}[!ht]
    \centering
    \includegraphics[scale=0.54]{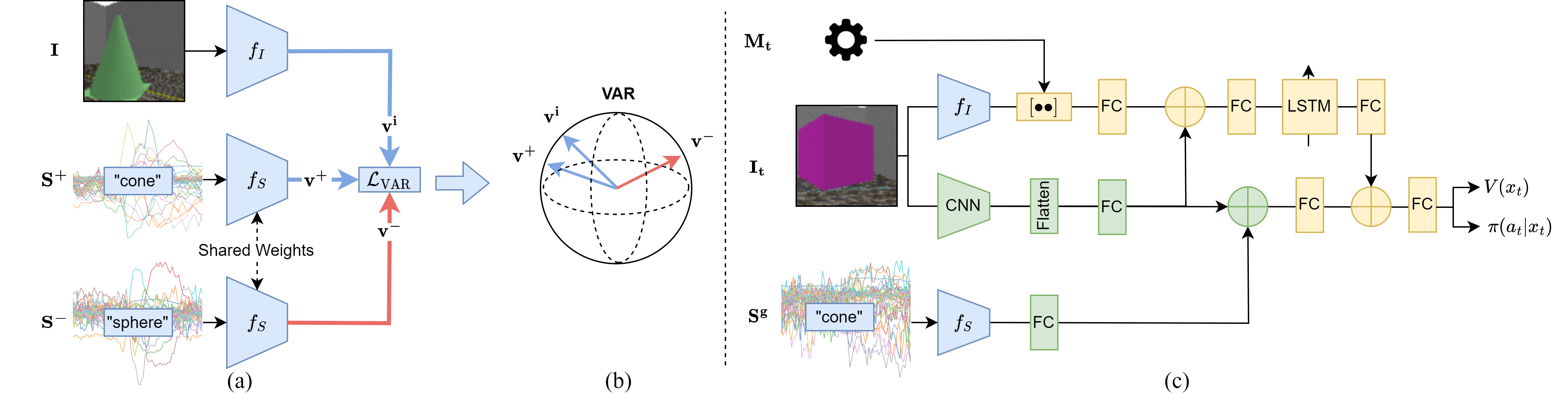}
    \caption{\textbf{The network architectures}. (a) The VAR is a three-branch Siamese network optimized with triplet loss. (b) The latent space of the VAR is a unit hypersphere such that vector embeddings of images and  audios of the same object are closer than those of  different objects. (c) The policy network for RL training. The portion in blue is VAR and the weights are frozen in RL training. We use $\bigoplus$ to denote element-wise addition, $FC$ to denote fully connected layers, and $[\bullet \bullet]$ to denote concatenation.}
    \label{fig:wholeNet}
    \vspace{-20pt}
\end{figure*}

\section{Methodology}
\label{sec:methods}
In this section, we describe our two-stage training pipeline and fine-tuning procedure. In training, we assume the availability of sufficiently large datasets and simulators. However, in fine-tuning, these resources are not available. 

\subsection{Visual-audio representation learning}
\label{sec:var}
In the first stage, we collect a dataset and learn a joint representation of images and audio, named VAR, that associates the image with its corresponding sound command. 

\subsubsection{Simulation environment and data collection}\label{sec:VAR_sim_data}
We let the robot randomly move in the environment and automatically collect visual-audio triplets $(\mathbf{I}, \mathbf{S^+}, \mathbf{S^-})$, where $\mathbf{I} \in \mathbb{R}^{n\times n \times 3}$ is the current RGB image from the robot's camera, $\mathbf{S^+} \in \mathbb{R}^{l\times m}$ is the Mel Frequency Cepstral Coefficients (MFCC)~\cite{davis1980comparison} of the positive sound commands, and $\mathbf{S^-} \in \mathbb{R}^{l\times m}$ is the MFCC of the negative sound commands. 
The positive and negative pair in a triplet tell the agent whether an image is semantically similar or dissimilar to an audio, while the exact class label of the image and the audio remains unknown to the agent. For example, when the TurtleBot sees a cone, it hears from the environment that the object is a cone $(\mathbf{S^+})$ but not a sphere $(\mathbf{S^-})$. 
When the TurtleBot does not see or is far away from all objects so that it sees many objects at once, it hears no positive sound, $\mathbf{S^+}=\mathbf{0}_{l\times m}$. 

\subsubsection{Training the VAR}Our goal is to encode both auditory and visual modalities into a joint latent space, where the embeddings of images and audios of the same intent are close together, and those of different intents are far apart. We formulate the problem as metric learning. As shown in Fig.~\ref{fig:wholeNet}a, 
the VAR is a Siamese network that has three branches for encoding the $\mathbf{I}$,  $\mathbf{S^+}$, and $\mathbf{S^-}$. We use $f_S: \mathbb{R}^{l\times m} \rightarrow \mathbb{R}^{d}$ to encode both $\mathbf{S^+}$ and $\mathbf{S^-}$, where $d$ is the dimension of the latent space, and we use $f_I: \mathbb{R}^{n\times n \times 3} \rightarrow \mathbb{R}^{d}$ to encode $\mathbf{I}$. 
Then, $\mathbf{v^i}=f_I(\mathbf{I})$, $\mathbf{v^+}=f_S(\mathbf{S^+})$ , and $\mathbf{v^-}=f_S(\mathbf{S^-})$ are the latent representation of $\mathbf{I}$, $\mathbf{S^+}$, and $\mathbf{S^-}$, respectively.
We enforce the norm of $\mathbf{v^i}$, $\mathbf{v^+}$, and $\mathbf{v^-}$ to be $1$ by applying a $L2$-normalization, such that the latent space lives on a unit hypersphere as shown in Fig.~\ref{fig:wholeNet}b. Practically, any deep models for sound and image processing can be used for $f_S$ and $f_I$. 
We use triplet loss as the objective, which encourages the distance between $\mathbf{v^i}$ and $\mathbf{v^-}$ to be greater than that between  $\mathbf{v^i}$ and $\mathbf{v^+}$ by a margin of $\alpha$~\cite{schroff2015facenet}:
\begin{equation}
\label{eq:L_VAR}
  \mathcal{L}_{\textrm{VAR}}=\max (0, \lVert \mathbf{v^i} - \mathbf{v^+} \rVert ^2_2 - 
  \lVert \mathbf{v^i}-\mathbf{v^-} \rVert ^2_2 + \alpha)
\end{equation}

\subsection{RL with visual-audio representation}
\label{sec:methods_rl}
The second stage of our pipeline is to learn a policy with RL using intrinsic reward functions from the trained VAR. 


\subsubsection{VAR-aided RL}
We model the interaction as a Markov Decision Process (MDP), defined by the tuple $ \langle \mathcal{X}, \mathcal{A}, P, R   \rangle$.  
At each time step $t$, the agent receives an image $\mathbf{I_t}$ from its camera and robot states $\mathbf{M_t}$ such as end-effector location or odometry. 
At $t=0$, the agent receives an additional one-time sound command $\mathbf{S^g}$ corresponding to an intent. We define the state $x_t \in \mathcal{X}$ for the MDP to be $x_t = [\mathbf{I_t}, f_I(\mathbf{I_t}), f_S(\mathbf{S^g}), \mathbf{M_t}]$, where $f_I$ and $f_S$ come from the VAR which is fixed in this stage of the pipeline. The VAR  compactly represents the high-dimensional observations and the intent contained in $\mathbf{S^g}$. Then, the agent takes an action $a_t\in\mathcal{A}$ based on its policy $\pi(a_t|x_t)$. In return, the agent receives a reward $r_t \in R$ and transitions to the next state $x_{t+1}$ according to an unknown state transition $P(\cdot|x_t, a_t)$. The process continues until $t$ exceeds the maximum episode length $T$ and the next episode starts.

\subsubsection{Intrinsic rewards}
Since the embeddings of an image and a sound command of the same intent are close to each other in the VAR, the intrinsic reward is the similarity between the current image and the sound command.
\begin{equation}
\label{eq:r_t}
  r^i_t= f_I(\mathbf{I_t}) \cdot f_S(\mathbf{S^g})
\end{equation}
Intuitively, the agent receives a high reward when the scene it sees matches the sound command. 
Chang \textit{et al} use extrinsic reward functions that rely on task-specific or robot-specific ground-truth information from the simulator~\cite{chang2020robot}. In contrast, our intrinsic reward function depends on the similarity of the embeddings and is thus agnostic to tasks and robots. 

RL agents trained with Eq.~\ref{eq:r_t} can already achieve decent performance. In reward calculation, providing the current sound signal $\mathbf{S_t}$ can further improve the performance. The $\mathbf{S_t}$ can be triggered in the same way as $\mathbf{S^+}$ in Sec.~\ref{sec:VAR_sim_data}:
\begin{equation}
\label{eq:r_t_current}
  r_t^i= f_I(\mathbf{I_t}) \cdot f_S(\mathbf{S^g}) + f_S(\mathbf{S_t}) \cdot f_S(\mathbf{S^g})
\end{equation}
The second dot product is the similarity between the current sound and the sound command. The $\mathbf{S_t}$ is not part of the state $x_t$ and is only used to calculate the reward. 
Thus, the robot policy is not conditioned on $\mathbf{S_t}$ and does not require sound feedback at test time.
Neither Eq.~\ref{eq:r_t} nor Eq.~\ref{eq:r_t_current} depends on any extrinsic task-specific reward. Nevertheless, we show that our model outperforms the baselines trained with extrinsic rewards in Sec.~\ref{sec:exp}.


\subsubsection{Network architecture}
Our policy network architecture is in Fig.~\ref{fig:wholeNet}c. The network takes $x_t$ as input and output the value $V(x_t)$ and the policy $\pi(a_t|x_t)$. The reason that we need another CNN in the policy network is that $f_I$ and the policy CNN may need to extract different features. For example, in a navigation task, the CNN needs to encode information about obstacles for collision avoidance, while $f_I$ does not need to. The embeddings of image $\mathbf{I_t}$ and robot state $\mathbf{M_t}$ are fed into an LSTM for long-term decision making, while the goal sound $\mathbf{S^g}$ is not passed into an LSTM because it is a one-time signal. We use Proximal Policy Optimization (PPO) for policy and value function learning \cite{schulman2017proximal}.


\subsection{Fine-tuning}
\label{sec:finetune}
 When the robot is deployed in the real world, our pipeline allows non-experts to fine-tune a trained model in a cheap and intuitive way. We assume it is difficult for non-experts to provide accurate speech transcriptions, hardware setup, labels for detection and recognition, demonstrations, or accurate measurement for the reward functions.
  Fortunately, fine-tuning our method needs none of these: non-experts only need to provide triplets based on their common knowledge using their own voices, which are used to fine-tune the VAR as in Sec.~\ref{sec:var}. The non-experts do not need to know the total number of objects or the class index of each object. We use Eq.~\ref{eq:r_t} as the reward function if $\mathbf{S_t}$ are not available. The robot can then self-improve its policy network with the intrinsic reward function by randomly sampling a collected sound command as the goal. No additional human supervision other than the collected triplets is needed for RL training. If a user is able to provide $\mathbf{S_t}$, we can use Eq.~\ref{eq:r_t_current} to perform real-time voice-based supervision to the robot.

\section{Simulation experiments} 
\label{sec:exp}
In this section, we first describe the simulation environments, illustrated in Fig.~\ref{fig:qualitative_result} and supplementary video, and various sound dataset for the experiments. Then, we compare the performance and data efficiency of our pipeline with several baselines and ablation models.


\subsection{Robotic Environments}
The environments are implemented using PyBullet~\cite{coumans2019} and AI2-THOR~\cite{ai2thor}. The TurtleBot and the Kuka environments pose difficulties in fine motor control and have moderate difficulty for perception. The iTHOR environment is challenging in perception, while the control is discretized and simplified. In all environments, the perception mainly comes from a monocular uncalibrated RGB camera.

\subsubsection{TurtleBot} The TurtleBot environment contains a $4\text{m}^2$ arena and four objects: a cube, a sphere, a cone, and a cylinder. Each object is associated with an intent. The robot's goal is to navigate to the object mentioned in a command based on RGB images. 
The agent needs to learn exploration skills to find the target object as quickly as possible.

\subsubsection{Kuka} The Kuka environment contains four identical blocks placed randomly in a line at a location unknown to the robot. 
Each block is associated with an intent. The robot's goal is to move its gripper tip to the top of the target block based on images from a location-fixed camera. 
The agent needs to move by observing its gripper tips and the block from RGB images affected by perspective distortion.

\subsubsection{iTHOR}
Our iTHOR environment is the first AI environment with real multi-word speech commands. The environment consists of 30 different floor plans of living rooms and simulates real-world applications of household robots. 
The intents include toggling on and toggling off the floor lamp and the television in a scene. The goal of the robot is to navigate to and manipulate the object mentioned in a sound command based on images and a noisy local collision map. 
The agent in the iTHOR environment needs to associate complex speech commands with rich visual observations and perform mapless navigation to finish the task.

\subsection{Sound data}
We consider various types of sounds from state-of-the-art datasets. We use speech signals from  Google Speech Commands (GSC)~\cite{warden2018speech} and Fluent Speech Commands (FSC)~\cite{lugosch2019speech} collected for contemporary end-to-end SLU. We use environmental sounds from UrbanSound8K (US8K)~\cite{Salamon:UrbanSound:ACMMM:14}, and single-tone signals from NSynth~\cite{nsynth2017}.

\begin{table}[h]
\vspace{-5pt}
  \begin{center}
    \caption{Sound signals used in the experiments.}
    \label{tab:sound}
    \begin{tabular}{@{}lll@{}} 
    \toprule
    
     \multirow{1}{*}{\textbf{Dataset}} &
     \multirow{1}{*}{\textbf{Sound}} &
     \multirow{1}{*}{\textbf{Examples}} \\
     \midrule
      
     \multirow{4}{*}{FSC}
      & activate light & ``Switch the lights on,'' ``Lamp on''\\
      & deactivate light & ``Turn the lamp off,'' ``Lights off''\\
      & activate music & ``Put on the music,'' ``Play''\\
      & deactivate music & ``Pause music,'' ``Stop''\\
      \midrule
      
      \multirow{2}{*}{GSC}
      & ``0,'' ``1,'' ``2,'' ``3'' & ``zero,'' ``one,'' ``two,'' ``three''\\
      & names of 4 objects & ``house'' ``tree,'' ``bird,'' ``dog''\\
      
      \midrule
      
      NSynth & $C_4$, $D_4$, $E_4$, $F_4$ & Various instruments, tempo, and volume \\
      
      US8K & bark, jackhammer & Sound recorded in the wild\\
      \bottomrule
    \end{tabular}
  \end{center}
  \vspace{-15pt}
\end{table}

We define Wordset to be the ``0,'' ``1,'' ``2,'' ``3'' in GSC. To show that our model can map different types of sounds to one object or concept, we mix ``house'' with ``jackhammer'' and ``dog'' with bark to form a \textit{Mix} dataset.

\subsection{Experiment setup}
We now introduce the training and evaluation of our pipeline in simulation. In our experiments, the overall architectures are kept the same for solving all the environments. 
\subsubsection{Training}
For all tasks, we collect 50,000 audio-visual triplets for the VAR training. 
The triplet margin $\alpha \in [1.0, 1.2]$ and the latent dimension $d$ is $3$. During RL training, the VAR is fixed and $8$ instances of the environment run in parallel. 

\subsubsection{Baselines and Ablations} 
We compare our method with three baselines and one ablation. The first baseline, denoted as ``RSI,'' is the representative of end-to-end language understanding agents which has the access to hand-tuned task-specific reward functions and ground-truth class labels for the images and sounds~\cite{chang2020robot}. The method uses auxiliary losses for image and audio classification during its RL training.

The second baseline, denoted as ``NoAux+SparseR,'' is a variant of RSI. We first learn image and audio classifiers using 50,000 one-hot labels with supervised learning. Then, we initialize RSI with the weights from the classifiers. During the RL training, we update the policy network without auxiliary losses, and we give the agent a reward of $1$ if a task is completed and $0$ otherwise at each timestep. 
This baseline shows the performance of ``RSI'' without ground-truth class labels and hand-tuned reward functions.

The third baseline, denoted as ``ASR+NLU+RSI,'' is a conventional pipeline. The speech is first transcribed by an off-the-shelf deep learning-based ASR named Mozilla DeepSpeech~\cite{hannun2014deep}. The output of the ASR could be noisy. For example, ``Play the music'' is sometimes transcribed as ``by the music.'' We therefore train a learning-based NLU using ground-truth transcription. The NLU is robust to the noisy text input and can output the correct intent. The intent is then transformed into an one-hot encoding, which is fed to the RSI architecture without a sound processing network. Notice that, in contrast, our method never uses transcriptions.

The ablation model, denoted as ``No current sound,'' uses Eq.~\ref{eq:r_t} as the reward function. The purpose of the ablation model is to quantify the effect of the current sound signal. 

\subsubsection{Definition of labels}
Labels include class labels and queries to the simulator. For example, one-hot labels for image and sound classification are class labels. Measuring the distance between the robot and the goal is a simulator query. One triplet requires two labels to indicate the positive and the negative. Every RSI training step requires 3 labels, including target object state checking and distance measuring to calculate the extrinsic reward, and at least 1 one-hot label for auxiliary losses. For convenience purposes, we count the one-hot labels for all the auxiliary losses as one for RSI.

\subsubsection{Evaluation metrics}
We evaluate the model with two metrics: (1) Success rate, defined as the percentage of successful episodes in all test episodes. We test the learned policy for $50$ episodes for each intent ($200$ episodes in total). (2) The amount of labels used for the training. 
A model that is suitable for fine-tuning in the real world requires as few annotations as possible. 

\begin{table}[t!]
  \centering
    \caption{Testing results in the Kuka and the TurtleBot environments with different types of sounds.}
    
    \label{tab:kuka_tb_test}
    \begin{threeparttable}
    \begin{tabular}{@{}lllccc@{}} 
    \toprule

     \multirow{2}{*}{\textbf{Environment}} & 
     \multirow{2}{*}{\textbf{Dataset}} &
     \phantom{abc} &
     \multicolumn{3}{c}{\textbf{Success rate}$\uparrow$} \\
     \cmidrule{4-6}
     &  && NoAux+SparseR & RSI & Ours \\ 
     \midrule
     \multirow{3}{*}{Kuka\tnote{a}} 
      & Wordset && 46.0 & 95.5 & \textbf{97.0} \\
      & NSynth && 28.5 & 92.5 & \textbf{98.0}\\
      & Mix && 23.0 & 94.0 & \textbf{95.5}\\ 
     \midrule
      
     \multirow{3}{*}{TurtleBot\tnote{b}} 
     & Wordset && 22.0 & 92.0 & \textbf{94.0} \\
      & NSynth && 9.5 & 95.0 & \textbf{96.0}\\
      & Mix && 21.0 & 87.0 & \textbf{91.0} \\ 
   
      \bottomrule
    \end{tabular}
    \begin{tablenotes}
   \item[a] Success if the gripper stays on the target block at least $50$ timesteps. 
   \item[b] Success if the robot stays in front of the target at least $5$ timesteps. 
  \end{tablenotes}
  \end{threeparttable}
    

  \vspace{-8pt}
\end{table}

\begin{table}[t!]
  \begin{center}
    \caption{Testing results in the iTHOR 201-220}
    \label{tab:iTHOR_test}
    \begin{threeparttable}
    \begin{tabular}{@{}llllc@{}} 
    \toprule
    
     \multirow{1}{*}{\textbf{Dataset}} &
     \multirow{1}{*}{\textbf{Models}} &
     \phantom{abc} &
     \multirow{1}{*}{\textbf{Label usage$\downarrow$}} &
     \multirow{1}{*}{\textbf{Success rate$\uparrow$}} \\
     \midrule
     
      \multirow{5}{*}{FSC}
       & NoAux+SparseR && $6.0 \times 10^6$ & 26.0 \\
      & No current sound && $\mathbf{0.1 \times 10^6}$ & 61.5 \\
      & ASR+NLU+RSI && $18.0 \times 10^6$ & 66.0 \\ 
      & RSI && $18.05 \times 10^6$ & 68.0 \\
      & Ours && $6.1 \times 10^6$ & \textbf{71.0} \\

      \bottomrule
     
    \end{tabular}
    
    \begin{tablenotes}
   \item[*] Success if the robot correctly toggles the floor lamp or television. 
    
  \end{tablenotes}
  \end{threeparttable}
  
  \end{center}
  \vspace{-25pt}
\end{table}

\subsection{Results for control policies}

\subsubsection{Unheard sounds}
\label{sec:unheardSounds}
In this experiment, all models are first trained with the same amount of RL steps. Then, we test the models with sound commands that never appear during training (e.g. new speakers)  without fine-tuning. We use this experiment to compare the performance of the models when the amount of data and the training steps are sufficient. For the iTHOR environment, the total training steps is 6 million (M), and the agent is tested within the training floor plans (Floor Plan 201 - 220). The total training steps is 3M for the other two environments.

From Table~\ref{tab:kuka_tb_test} and Table~\ref{tab:iTHOR_test}, we find that our method generalizes well in terms of different types of sound commands and robotic tasks. Our method achieves $1\% \sim 5.5\%$ higher success rates compared with RSI in all three environments, even though it has no access to ground-truth class labels or extrinsic reward functions during the whole RL training. 
We believe the reason is that RSI optimizes multiple losses jointly, which may cause biased gradients. In contrast, our two-stage pipeline learns the two tasks separately, which benefits the training of both tasks. 
The success rates for ``NoAux+SparseR'' baseline for all the environments are much lower than RSI and our method. The result suggests that the performance of RSI degrades significantly without unlimited ground-truth labels for its auxiliary losses and heavily hand-engineered reward function. In contrast, our method can perform better even with weaker supervision and fewer engineering efforts and is thus more general.

From Table~\ref{tab:iTHOR_test}, the success rate for ``ASR+NLU+RSI'' baseline is lower than both the RSI and our method, due to intermediate errors, as mentioned in Sec.~\ref{sec:relatedA}. This result justifies the usage of end-to-end SLU in RSI and our work. The success rate for ``No current sound'' ablation is only $9.5\%$ lower than our method, but the label usage of this ablation is significantly lower than all other models. The result indicates that current sound signals benefit the performance. However, our method works reasonably well without the current sound signals.


\subsubsection{Fine-tuning}
We use this sim2sim experiment to show the label efficiency of our method when the robot is fine-tuned in a new domain. For example, a trained household robot needs to serve and navigate in an unseen living room.
A high-efficiency method should achieve a higher success rate than other methods given the same amount of new labels.

We first directly test the performance of trained models with unheard sound commands in each unseen iTHOR floor plan. No fine-tuning is performed and $0$ new label is collected. From the first two columns of Table~\ref{tab:iTHOR_finetune_small}, we see that the performance drops for both methods due to domain shift, a common phenomenon in learning systems, which suggests the necessity of fine-tuning. We then collect 5000 new labels in one unseen floor plan to fine-tune the models for that floor plan. For our method, we follow Sec.~\ref{sec:finetune} to fine-tune the VAR with new triplets and use Eq.~\ref{eq:r_t} to self-improve RL policy without current sounds. For RSI, we collect one-hot labels and use simulator queries during its RL training. 
 

    
    
     
     
      
      
     
     

\begin{table}[t]
  \begin{center}
    \caption{Average success rates over unseen iTHOR Floor Plan 226 - 230 after fine-tuning.}
    \label{tab:iTHOR_finetune_small}
    \begin{tabular}{@{}llllllll@{}}
    \toprule
    
     \textbf{New labels} & \multicolumn{2}{c}{0} & \multicolumn{4}{c}{5000} \\ 
     \cmidrule{1-1}
     \cmidrule(lr){2-3}
    \cmidrule(lr){4-7} 

     \multirow{2}{*}{\textbf{Methods}} & \multirow{2}{*}{RSI} &\multirow{2}{*}{Ours} & \multirow{2}{*}{RSI} & Ours& Ours  & Ours \\
     & & & & 0.08M & 0.4M & 1M \\ 
     \midrule
     \textbf{Avg.$\uparrow$} & 30.7 & 31.5 & 37.2 & 67.5 & 73.7 & \textbf{84.7} \\

      \bottomrule
    \end{tabular}
  \end{center}
  \vspace{-25pt}
\end{table}

From Table~\ref{tab:iTHOR_finetune_small}, we find that the RSI baseline can only be improved by $6.5\%$ using 5000 labels, while our method improves itself by  $36\%$, $42\%$, and $53\%$ after $0.08$M, $0.4$M, and $1$M RL training steps, respectively, using the same amount of labels. 
In the fine-tuning stage, RSI uses the labels for policy network fine-tuning, which leads to fast depletion of the label quotas and less RL experience. The result suggests that fine-tuning the RSI is unrealistic. The richer RL experience in our method is due to its higher efficiency because labels are used to update the intrinsic reward and there is no label consumption during its self-supervised RL exploration. 
Therefore, in new domains where the number of labels are constrained, our method has the potential to achieve higher performance with a less laborious fine-tuning than RSI.
\begin{figure}[h]
    \centering
    
    \includegraphics[scale=0.13]{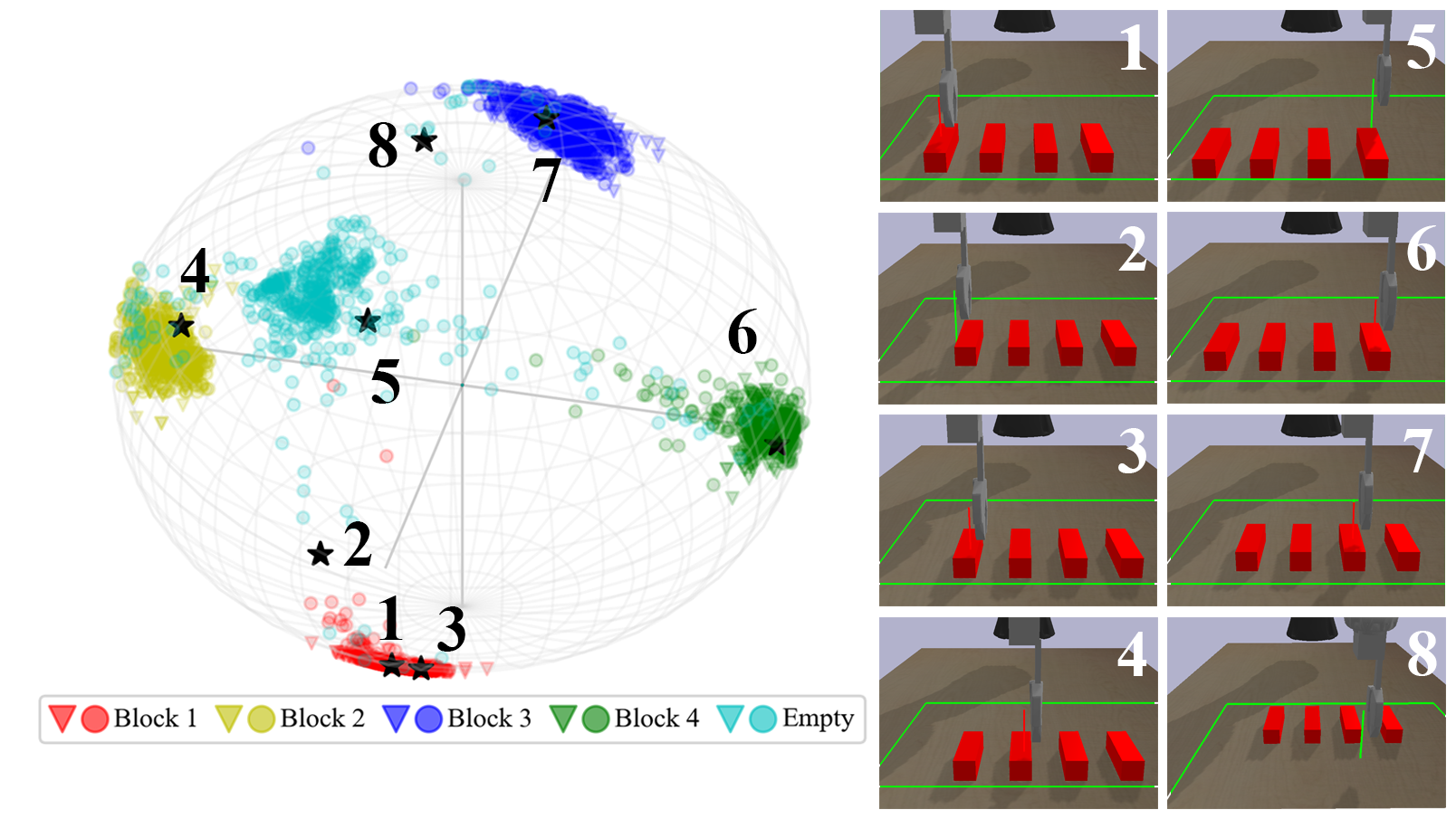}

    \caption{\textbf{Visualizations of the VAR in the Kuka environments with Wordset.} The colors indicate the ground truth labels of sound and image data. The embeddings of images and sound are marked by circles and triangles, respectively. The black stars are the vector embeddings of the 8 images. Block 1 is the leftmost block. The ``Empty'' class consists of empty sounds and images of the gripper tip above none of the blocks.}
    \label{fig:rep}
     \vspace{-10pt}
\end{figure}

\begin{figure}[h]
    \centering
    \includegraphics[scale=0.15]{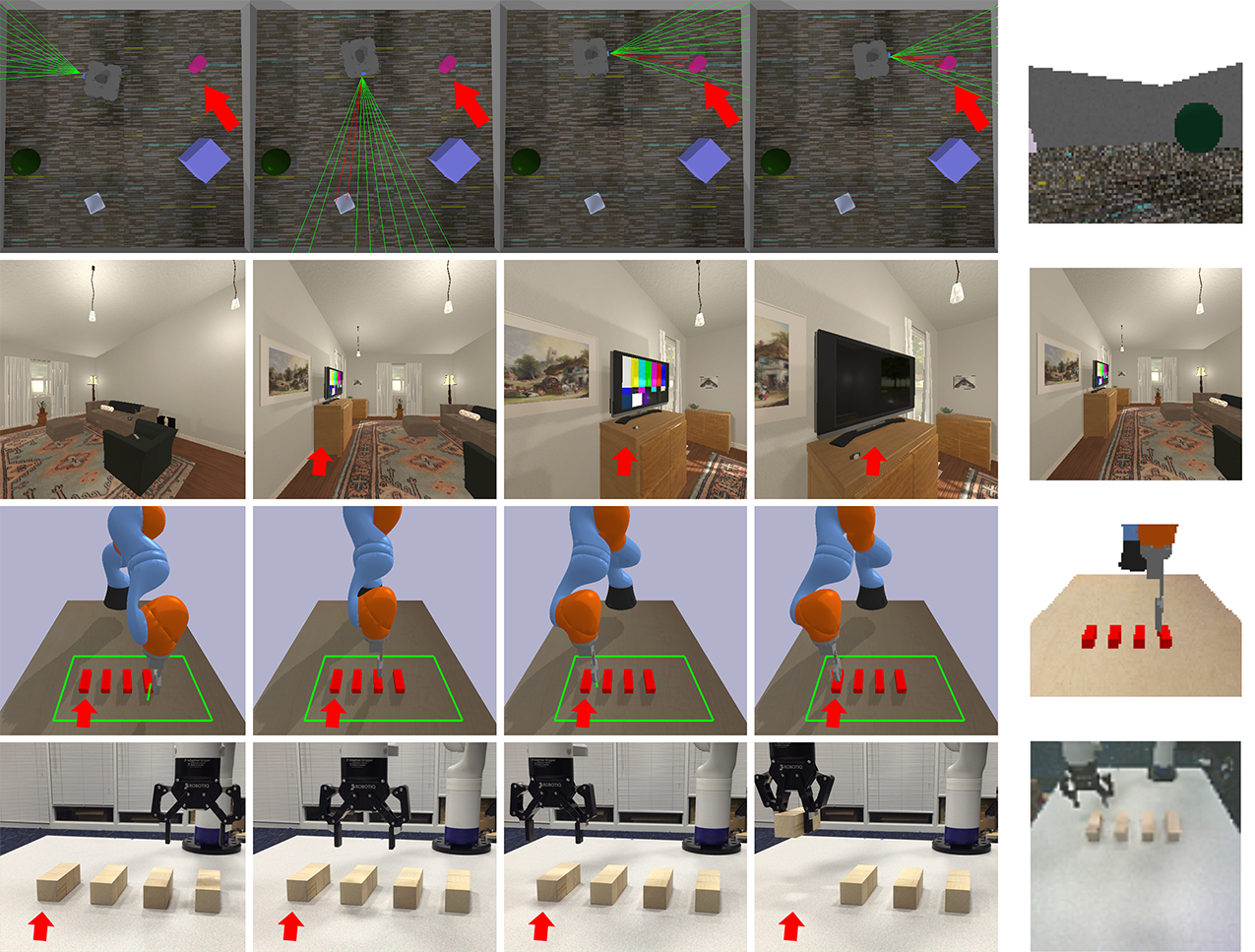}
    \caption{\textbf{Qualitative Result and environment illustration}. The target objects are indicated by the red arrows. Green and red lines are invisible to the robots. Right-most column shows the robot camera view. \textit{Top}: Simulated TurtleBot searches and approaches its target, the purple cylinder. \textit{Middle}: iTHOR agent searches for the television and turn it off. \textit{Bottom}: The simulated Kuka and a real Kinova Gen3 move grippers to the target block.}
    \label{fig:qualitative_result}
    \vspace{-10pt}
\end{figure}



     
     
      



\subsubsection{Qualitative results}
Fig.~\ref{fig:qualitative_result} and the attached video show examples of the agent successfully reaching their goals. 

\subsection{Qualitative results of the VAR}
To better understand the VAR and how it produces an effective intrinsic reward, we project $\mathbf{I}$ and  $\mathbf{S^+}$ from the collected triplets using a trained VAR, as shown in Fig.~\ref{fig:rep}. We see that $\mathbf{v^i}$ and $\mathbf{v^+}$ of the same concept form a cluster and all clusters are separated from each other. Thus, the agents successfully associate images and sounds with the same meanings and distinguish those with different meanings,  even though we do not explicitly classify the objects or inform the agent of the number of classes. 

We also find that the VAR is able to map the images with ambiguous labels to meaningful locations on the spheres. For example, in Fig.~\ref{fig:rep}, the gripper tip in \#2 is not directly above but is very close to Block 1. In the VAR sphere, the vector embedding is indeed between the Block 1 (red) and the Empty (cyan) cluster. 
With the ability to represent ambiguous images, the VAR generates a relatively dense reward function that guides the robot exploration in RL. 


\section{Real-world experiments}
\label{sec:real_exp}
We use a real Kinova Gen3 arm for the Kuka choosing task to show the effectiveness of our method in the real world. We choose this task for the experiment because the performance of the models easily degrade due to the inevitable inconsistency in the camera pose between simulation and the real world. 
We compare our method with domain randomization, a widely used strategy for sim2real~\cite{Tobin_2017} and show the advantage of our real-world fine-tuning. We use Wordset as sound commands. 
The agent controls the robot gripper to move towards the target block and adjusts its pre-grasp pose. At the end of an episode, the robot lowers its gripper vertically and perform a grasp. 
\begin{table}[h]
 \vspace{-5pt}
  \begin{center}
    \caption{Success counts for each intent and average success rate with a total of 48 tests (12 for each intent)}
    \label{tab:real_exp}
    \begin{tabular}{llllll} 
    \toprule
     \textbf{Words} & ``0'' & ``1'' 
     & ``2'' & ``3'' & avg.$\uparrow$ \\
      \midrule
    \textbf{Domain randomization} & 8 & 11 & 9 & 7 & 72.9\\
      \textbf{Fine-tuning} & 11 & 9 & 11 & 12 & \textbf{89.6}\\
      \bottomrule
    \end{tabular}
  \end{center}
  \vspace{-15pt}
\end{table}

\textbf{Results with domain randomization}:
We randomize relative positions and texture of the blocks, camera viewpoint, and background. In the first row of Table~\ref{tab:real_exp}, we see that the model with domain randomization only helps bridge the sim2real gap to some extent, but still insufficient for real-world learning-based robotic applications. 

\textbf{Results with fine-tuning}:
 Our fine-tuning further diminishes the sim2real gap. We spend about only 1 hour in manually collecting 993 triplets and another hour to fine-tune the VAR and the RL policy for 10000 steps. No additional labels or supervision are needed from the humans. In Table~\ref{tab:real_exp}, the average performance after fine-tuning increases $16.7\%$ compared with domain randomization, which is significant given the relatively small number of real-world data and training time. In the failure cases, the agent only fails at grasping because the block is not completely within its jaw. A typical successful episode is shown in Fig.~\ref{fig:qualitative_result}. 

\section{Conclusion and Future work}
\label{sec:conclusion}

We propose a general and self-improving pipeline for vision-based voice-controlled robots. Our pipeline learns a VAR which generates an intrinsic reward function and embeddings for the downstream control tasks. 
Our method requires fewer labels yet outperforms baselines that have access to extrinsic reward functions. This gain extends across diverse robotic tasks in simulation. We also successfully transfer the policy to a real robot and improve the policy data-efficiently in the real world. 
Possible directions for the future work include: (1) generalizing the same VAR to different downstream tasks, and (2) learning the representation without requiring negative pairs.





\newpage
\clearpage
\bibliographystyle{IEEEtran}
\bibliography{BibFile}
\clearpage
\end{document}